# ROI SEGMENTATION FOR FEATURE EXTRACTION FROM HUMAN FACIAL IMAGES

Surbhi[1], Vishal Arora[2]

[1]M.Tech CSE, Shaheed Bhagat Singh College of Engineering & Technology, Ferozepur, Punjab

Email: surbhibatta@gmail.com

[2]Assistant Professor, Shaheed Bhagat Singh College of Engineering & Technology, Ferozepur, Punjab

Email: vishal.fzr@gmail.com

*Abstract:* Human Computer Interaction (HCI) is the biggest goal of computer vision researchers. Features form the different facial images are able to provide a very deep knowledge about the activities performed by the different facial movements. In this paper we presented a technique for feature extraction from various regions of interest with the help of Skin color segmentation technique, Thresholding, knowledge based technique for face recognition.

## I. INTRODUCTION

Segmentation is very important to image retrieval process. The shape feature and the layout feature both depend on good segmentation technique. Image segmentation algorithms are generally based on one of the two basic properties of intensity values: *Discontinuity* and *Similarity*. In the first category, the approach is to partition an image based on abrupt changes in intensity, such as edges in an image. The principal approaches in the second category are based on partitioning an image into regions that are similar according to a set of predefined criteria. Thresholding, Region growing, Region splitting and Merging are examples of such methods in this category. In Region based segmentation the objective is to partition an image into regions. This paper explains the algorithms for finding Region of Interest from the facial images and extraction of features from the respective Region of Interest. In the proposed techniques, the different ROI's from the facial data are taken as Lips and Eyes of human facial image. These are detected from the face with the help of skin color and the knowledge based methods in concern with the human facial data.

## II. FEATURE EXTRACTION

For the extraction of features from the various Regions of Interests, we draw the curves in these regions by considering those pixels, which are not considered as skin pixels in the respective regions. After the detection of non-skin pixels we apply the flood fill algorithm to find the curves from the non-skin pixels. The flood fill algorithm takes three parameters: a start node, a target color, and a replacement color. The algorithm looks for all nodes in the array which are connected to the start node by a path of the target color, and changes them to the replacement color. Figure 1 shows the implementation of Flood-fill algorithm:

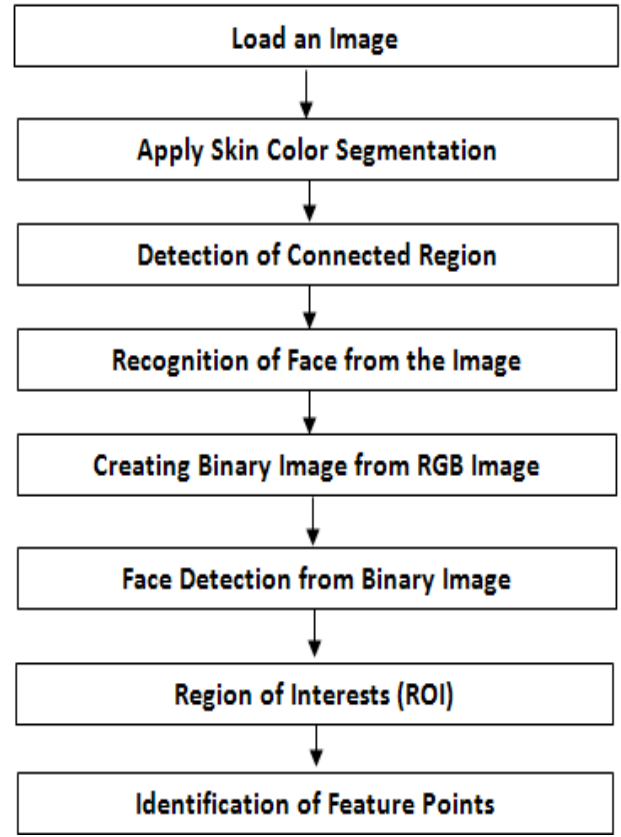

*Figure 1: Methodology Used in proposed work*

**Flood**-fill (*node, target-color, replacement-color*)

1. Set Q to the empty queue.
2. If the color of node is not equal to target-color, return.
3. Add node to Q.

4. For each element n of Q:
5. If the color of n is equal to target-color:
6. Set w and e equal to n.
7. Move w to the west until the color of the node to the west of w no longer matches target-color.
8. Move e to the east until the color of the node to the east of e no longer matches target-color.
9. Set the color of nodes between w and e to replacement-color.
10. For each node n between w and e:
11. If the color of the node to the north of n is target-color, add that node to Q.
12. If the color of the node to the south of n is target-color, add that node to Q.
13. Continue looping until Q is exhausted.
14. *Return.*

After getting the curves for left eye, right eye and lip regions. We consider the extreme points of these curves and first two feature points and other feature points are determined by drawing tangents on curves found from these three different regions. The proposed algorithm gives better results for determination of ROI's and for obtaining feature points.

### III. OBJECTIVES

- To recognize face from the image supplied.
- To identify ROI to work with different parts of facial image.
- To find the shape curves from each ROI.
- To perform Feature extraction with the help of tangents to the extreme feature points.

### IV. METHODOLOGY

Step 1. Skin Color Segmentation algorithm is applied to the loaded image to find the skin color for the detection of face from the image given to the system. Skin color is found from the image with the help of threshold as:

$$if \frac{R + G + B}{3} \geq T \quad \begin{cases} P(x,y) = white \\ P(x,y) = black \end{cases}$$

where $P(x, y)$, represents the current pixel with co-ordinate positions x and y. And R, G, B represents the Red, Green and Blue values associated with current pixel $P(x, y)$ and T is the threshold value for skin color pixels.

Step 2. Maximum width and Maximum heights of connected skin pixels are considered as the width and height of the face present in the image.

Step 3. To check the possibility to have a face in the given image, the height and width of the face area must follow the following criteria:

$$height \geq 50, width \geq 50$$

$$1 \leq \frac{height}{width} \leq 2$$

Step 4. After Face Recognition, convert the facial image into Binary image.

Step 5. Then cut the Face from the binary image according to skin color pixels present in the image by considering height and width.

Step 6. Then find Region of Interests (ROI) from the face as Left Eye, Right Eye and Lip according to knowledge based method of face detection.

Step 7. Then apply Flood Fill Algorithm on Left Eye, Right Eye and Lip are to find their Curves.

Step 8. Now consider the two extreme points of Curve as first two feature points of each region.

Step 9. Now by drawing tangents on the curves we pick four feature points on each Region of Interest.

### V. SIGNIFICANCE OF PROPOSED WORK

Feature based method shows efficient results as expressions are not checked with the help of all the pixels present in the image but this method just finds some specific points in the entire region according to the area or region of interest. These specific points are called the Feature points. To obtain these feature points firstly, we have to find the face from the facial image data. Secondly, from the specific regions of the face we have to predict the position of these feature points.

### VI. PERFORMANCE MEASUREMENT

Performance is measured in terms of the accuracy of face recognition phase, and accuracy of ROI detection from various facial images.

#### A. Accuracy of Face Recognition

To evaluate the performance, proposed face recognition algorithm is tested by supplying number of facial images N and number of Non Facial Images M. And then check the accurate rate of detection as:

$$A_f = \left(\frac{I_f}{N} * 100\right)$$

$$A_{nf} = \left(\frac{I_{nf}}{M} * 100\right)$$

Where $A_f$ is the percentage of detection accuracy achieved by facial image, $I_f$ is the total number of facial images correctly recognized by the proposed face recognition system. And $A_{nf}$ is the percentage accuracy achieved by Non facial images; $I_{nf}$ is the total number of non-facial images correctly recognized by the proposed face recognition system.

*B. Accuracy of ROI detection*

To evaluate the performance, proposed Region of Interests algorithm is tested by supplying number of facial images N. And then check the accurate rate of detection of ROI as:

$$A_{lip} = \left(\frac{ROI_{lip}}{N} * 100\right)$$

$$A_{l\_eye} = \left(\frac{ROI_{l\_eye}}{N} * 100\right)$$

$$A_{r\_eye} = \left(\frac{ROI_{r\_eye}}{N} * 100\right)$$

Where $A_{lip}, A_{l\_eye}, A_{r\_eye}$ are the percentage detection rate of exact Regions of interests for Lip area, Left Eye area and Right Eye area respectively; $ROI_{lip}, ROI_{l\_eye}, ROI_{r\_eye}$ are the total number of time Lip region found, Left Eye region found and Right Eye region found respectively and N is the total number of facial images applied to the Proposed ROI detection system.

## VII. CONCLUSION

Segmentation technique is quite effective to give some knowledge about the hidden properties of any image. As in the case of the present paper, skin color segmentation is used to identify or recognize the human face from the image and then the knowledge based method gives the idea about the different regions of the face in a facial image. After that the feature points are detected by drawing tangents on the curves obtained from different facial region of interests. The combination of these technologies must be proved good for these type of problem.

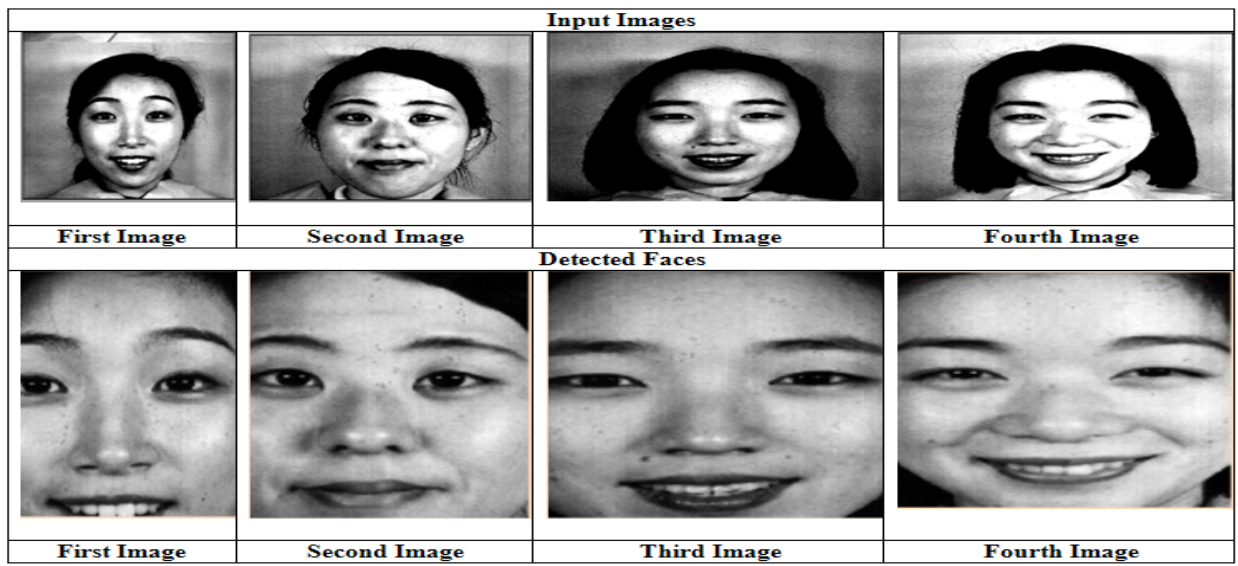

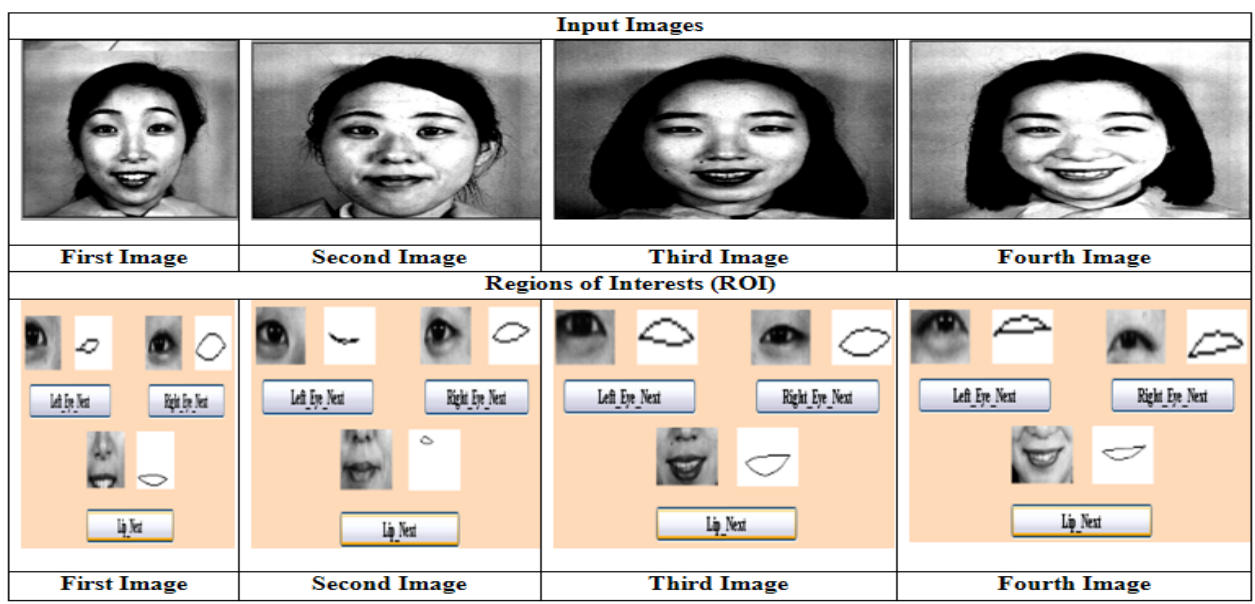